\documentclass[sigconf]{acmart}

\AtBeginDocument{%
  }

\usepackage{adjustbox}     

\usepackage{microtype}
\usepackage{graphicx}
\usepackage{amsmath}
\usepackage{lipsum}
\usepackage{algorithm}
\usepackage{algpseudocode}
\usepackage{amsfonts}
\usepackage{subcaption}
\usepackage{comment}
\usepackage{url}
\usepackage{multirow}
\usepackage{multirow}
\usepackage{booktabs}
\DeclareUnicodeCharacter{2212}{-}
\usepackage{makecell}
\usepackage{fdsymbol} %show footnote in number
\usepackage[T1]{fontenc}
\usepackage[utf8]{inputenc}

\usepackage{microtype}

\copyrightyear{2023}
\acmYear{2023}
\setcopyright{rightsretained}
\acmConference[AIES '23]{AAAI/ACM Conference on AI, Ethics, and Society}{August 8--10, 2023}{Montréal, QC, Canada}
\acmBooktitle{AAAI/ACM Conference on AI, Ethics, and Society (AIES '23), August 8--10, 2023, Montréal, QC, Canada}
\acmDOI{10.1145/3600211.3604665}
\acmISBN{979-8-4007-0231-0/23/08}

\begin{document}


% \title{Analysis of Climate Campaigns on Facebook using Model Soup Approach} 
% \title{Predicting Stance of Climate Campaigns in \\ Social Media using Model Soup Approach}
% \title{Predicting Stance of Climate Campaigns in \\ Social Media using Bayesian Model Averaging}
\title{Analysis of Climate Campaigns on Social Media \\ using Bayesian Model Averaging}

\author{Tunazzina Islam}
\affiliation{%
  \institution{Department of Computer Science, Purdue University}
  %\streetaddress{8600 Datapoint Drive}
  \city{West Lafayette}
  \state{IN-47907}
  \postcode{47907}
  \country{USA}
  }
\email{islam32@purdue.edu}

\author{Ruqi Zhang}
\affiliation{%
  \institution{Department of Computer Science, Purdue University}
  %\streetaddress{8600 Datapoint Drive}
  \city{West Lafayette}
  \state{IN-47907}
  \postcode{47907}
  \country{USA}
  }
\email{ruqiz@purdue.edu}

\author{Dan Goldwasser}
\affiliation{%
  \institution{Department of Computer Science, Purdue University}
  %\streetaddress{8600 Datapoint Drive}
  \city{West Lafayette}
  \state{IN-47907}
  \postcode{47907}
  \country{USA}
  }
\email{dgoldwas@purdue.edu}

%\author{Anonymous Submission}

\begin{abstract}
  Climate change is the defining issue of our time, and we are at a defining moment. Various interest groups, social movement organizations, and individuals engage in collective action on this issue on social media. In addition, issue advocacy campaigns on social media often arise in response to ongoing societal concerns, especially those faced by energy industries. Our goal in this paper is to analyze how those industries, their advocacy group, and climate advocacy group use social media to influence the narrative on climate change. In this work, we propose a minimally supervised model soup \cite{wortsman2022model} approach combined with messaging themes to identify the stances of climate ads on Facebook. Finally, we release our stance dataset, model, and set of themes related to climate campaigns for future work on opinion mining and the automatic detection of climate change stances. 
  % Our data, code, and model are publicly available %\href{https://drive.google.com/drive/folders/1nCdSfcByYEUAap_ZbmdjnZkoCxAjyywQ?usp=sharing}
  % \href{https://github.com/tunazislam/BMA-FB-ad-Climate}{here}.
\end{abstract}

\begin{CCSXML}
<ccs2012>
   <concept>
       <concept_id>10010147.10010178.10010179</concept_id>
       <concept_desc>Computing methodologies~Natural language processing</concept_desc>
       <concept_significance>300</concept_significance>
       </concept>
   <concept>
       <concept_id>10002951.10003260.10003272.10003273</concept_id>
       <concept_desc>Information systems~Sponsored search advertising</concept_desc>
       <concept_significance>500</concept_significance>
       </concept>
 </ccs2012>
\end{CCSXML}

\ccsdesc[300]{Computing methodologies~Natural language processing}
\ccsdesc[500]{Information systems~Sponsored search advertising}
\keywords{social media, climate campaigns, facebook ads, bayesian model averaging, minimal supervision}


\maketitle

\section{Introduction}
We are approaching a decisive moment for international efforts to tackle the climate crisis, and  \href{https://www.iea.org/}{International Energy Agency} (IEA) report sets out a pathway for achieving this goal by reducing global carbon dioxide ($CO_2$) emissions to \textit{net zero by 2050}. IEA emphasizes policy interventions by governments worldwide to drive the energy transition and lower greenhouse gas emissions. Towards a net-zero future, the United Nations (UN) campaign for individual action on climate change and sustainability called ActNow\footnote{\url{https://www.un.org/en/actnow}} so that by making choices that have less harmful effects on the environment, we can be part of the solution and influence change.   
Despite the urgency to avoid catastrophic climate change \cite{moritz2013future}, scientific explanation \cite{dessler1995science}, the policy plans of the world's governments \cite{adger2003adaptation}, digital activism \cite{hestres2017internet}, we are still lagging from climate goals. The reason behind this lag is the negative influence of fossil fuel companies working to undermine and weaken much-needed climate action \cite{nosek2020fossil}.

Over the last decade, online advertising has significantly increased to disseminate agendas and sponsored content has been used to reach more people on social media \cite{islam2023weakly,goldberg2021shifting,krupenkin2021vaccine,capozzi2021clandestino,burki2020online}. Advertising plays a pivotal role in climate change because some advertising defends the destructive oil and gas industry, greenwashes brands and drives consumption. At a congressional hearing in April 2021, Facebook chief Mark Zuckerberg admitted that climate misinformation was a ``big issue''\footnote{\href{https://www.theguardian.com/technology/2021/sep/16/facebook-climate-change-misinformation}{www.theguardian.com}}. A Bloomberg analysis pointed out that millions of climate change-denial ads continue to be approved on the platform despite increasing pressure from climate groups to more effectively regulate content\footnote{\href{https://www.campaignasia.com/article/are-social-platforms-complicit-in-climate-change-misinformation/476473}{www.campaignasia.com}}. Oil and gas industries have been using paid-for social media advertising on Facebook to capture the narrative on climate change. 
However, climate scientists have reached a consensus that climate change is real and is caused by human activity on the planet, which has and will have adverse effects on humanity and the biosphere around the planet \cite{cook2016consensus}. Stakeholders supporting climate change also use the Facebook advertising platform to influence the targeted audience, focusing on transitioning to renewable energy. Though the transition to a renewable energy economy may be exciting to renewable energy advocates and scholars, many industries and community has different perspectives on it \cite{morris2019risk,richardson2021supporting}. 
\begin{figure}[htbp]
  \centering  
  \includegraphics[width= 1\columnwidth]{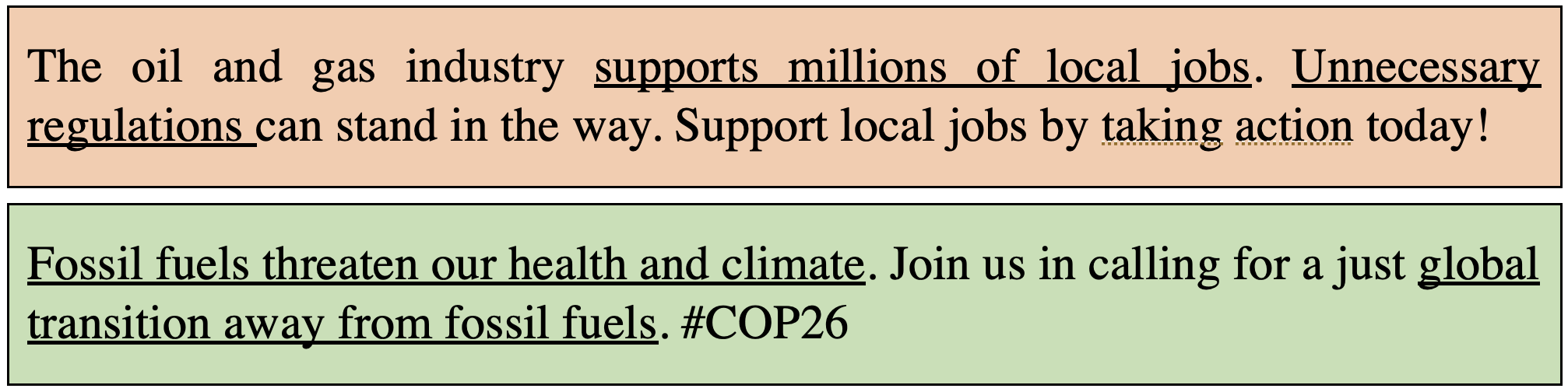}
    \caption{Example of sponsored ads in Facebook where the advertisers have different stances on climate change focusing on different themes.}
    \label{fig:climate_ads}
\end{figure}
For example, Fig. \ref{fig:climate_ads} presents two sponsored ads on Facebook having two different stances on climate change. The \textit{stance} of the top ad (inside the brown box in Fig. \ref{fig:climate_ads}) is ({\small\texttt{pro-energy}}) as the sponsor is against `unnecessary regulations on oil and gas industry' and the ad \textit{theme} is ({\small\texttt{economy\_pro}}) mentioning `oil and gas industry supports local jobs'. 
The \textit{stance} of the bottom ad (inside the green box in Fig. \ref{fig:climate_ads}) is ({\small\texttt{clean-energy}}) as the sponsor supports `transition away from fossil fuels' and the reason for this is the `threatening effect of fossil fuels on our health'. So the ad \textit{theme} is ({\small\texttt{HumanHealth}}).
% The \textit{stance} of the bottom ad (inside green box in Fig. \ref{fig:climate_ads}) is ({\small\texttt{clean-energy}}) as the sponsor supports `investment in renewable energy' and the ad \textit{theme} is ({\small\texttt{economy\_clean}}) mentioning `renewable energy creates good paying jobs'.

In this work, we aim to understand how climate advocates and fossil fuel corporations are using advertising to control the narrative on climate change and climate policy. 
%Our goal is twofold: first to identify the stances of the ads i.e., pro-energy vs clean-energy, and second to build on this identification to characterize the themes of the ads.
Our goal is twofold: first, to characterize the themes of the ads, and second to build on this characterization to identify the stances of the ads, i.e., \textbf{pro-energy}, \textbf{clean-energy}, \textbf{neutral}. 

Our theme assignment process is motivated by a thematic analysis approach \cite{braun2012thematic}. We begin by defining a seed set of relevant arguments based on recent studies \cite{miller2016audience,cha2021workers},
where each pro-energy theme is defined by multiple sentences. Since the initial set of themes contains only pro-energy arguments, we add clean-energy themes and phrases. We fine-tune a pre-trained textual inference model using a contrastive learning approach to identify paraphrases in a large collection of climate related ads.

In recent years, research has shown that models pre-trained
on large and diverse datasets learn representations that transfer well to a variety of tasks \cite{jiao2020tinybert,edunov2019pre,chronopoulou2019embarrassingly,howard2018universal}. The fine-tuning process has two steps: (1) fine-tune models with a variety of hyperparameter configurations, and (2) select the
model which achieves the highest accuracy on the held-out validation set and discard remaining models. \citet{wortsman2022model} recently showed that selecting a single model and discarding the rest has several downsides, and they proposed \textit{model soup}, which averages
the weights of fine-tuned models independently. While \citet{wortsman2022model} showed model soup performance on four text classification datasets from the GLUE benchmark \cite{wang2018glue}, we develop a minimally supervised model soup approach leveraging messaging theme to detect stance for analyzing climate campaigns on Facebook. 
%\ti{Need to add the summary of proposed method}
%
We focus on the following research questions (RQ) to analyze climate campaigns on social media:
\begin{itemize}
   \setlength\itemsep{.1 em}
    \item \textbf{RQ1.} Can a model trained with minimal supervision using theme information be leveraged to predict the presence of stances in Facebook ads related to climate change?
    \item \textbf{RQ2.} What are the intersecting themes of the messaging? 
    \item \textbf{RQ3.} What demographics and geographic are targeted by the advertisers?
    \item \textbf{RQ4.} Do the messages differ based on entity type?
\end{itemize}

Our contributions are summarized as follows:
\begin{enumerate}
    \item We formulate a novel problem of exploiting minimal supervision and Bayesian model averaging to analyze the landscape of climate advertising on social media.
    % \item We create a novel problem of exploiting minimal supervision and Bayesian model averaging to analyze the landscape of climate advertising on social media.
    \item We identify the themes of the climate campaigns using an unsupervised approach.
    \item We propose a minimally supervised model soup approach to identify stance combining themes of the content of climate campaigns. We show that our model outperforms the baselines. 
    \item We conduct quantitative and qualitative analysis on real-world dataset to demonstrate the effectiveness of our proposed model.
\end{enumerate}
% The rest of the paper is organized as follows: we start with
% the discussion of related work; next, we provide the dataset details; then, we show the problem
% formulation; next, we describe the methodology; later, we elaborate on details
% of experimental settings, including discussion of the results, baselines, ablation study; finally, we show the
% analysis by answering research questions specifically \textbf{RQ2}, \textbf{RQ3}, and \textbf{RQ4}. 
The remaining sections of the paper are structured as follows: we commence with a discussion on related work, followed by the presentation of dataset details. Subsequently, we introduce the problem formulation, after which we outline the methodology employed. Later, we provide comprehensive information on the experimental settings, including the results, baselines, and ablation study. Finally, we address the research questions \textbf{RQ2}, \textbf{RQ3}, and \textbf{RQ4} through a detailed analysis.
Our data, code, and model are publicly available at \url{https://github.com/tunazislam/BMA-FB-ad-Climate}

% %\href{https://drive.google.com/drive/folders/1nCdSfcByYEUAap_ZbmdjnZkoCxAjyywQ?usp=sharing}
%   \href{https://github.com/tunazislam/BMA-FB-ad-Climate}{here}.
\section{Related Work}
Recent studies have shown climate change activism in social media and news media \cite{bloomfield2019circulation,walter2018echo,stoddart2016canadian}. Sponsored content on social media -- especially Facebook, is the main channel to reach the targeted audience on a specific event such as US Presidential election \cite{islam2023weakly}, or specific issues, i.e., COVID \cite{islam2022covidfbAd,silva2021covid,mejova2020covid}, immigration\cite{capozzi2020facebook,ribeiro2019microtargeting}.
% We take a different approach to analyzing climate ads by identifying stances leveraging themes. 

Several studies have analyzed the discourse around climate change. \citet{luo2020detecting} proposed an opinion framing task on the global warming debate on media. \citet{koenecke2019learning} studied whether climate change related sentiment in tweets changed in response to five natural disasters occurring in the US in 2018. \citet{tweetstance} explored stance with respect to certain topics, including climate change in a tweet-based setting. To understand the narratives of climate change skepticism, \citet{bhatia2021automatic} studied the automatic classification of neutralization techniques. \citet{diggelmann2020climate} introduced a veracity prediction task in a fact-checking setting on climate claims. Our work differs from these in that we use a \textbf{probabilistic approach} to detect stance incorporating \textbf{theme information} of climate related ads on social media. 

Our work falls in the broad scope of minimal supervision \cite{islam2022covidfbAd,islam2022twitter,pacheco2022holistic,mekala2020meta,ratner2018snorkel,belkin2006manifold}, contrastive learning \cite{wang2021contrastive,gao2021simcse,giorgi2020declutr,wu2020clear} and Bayesian model averaging \cite{madigan1996bayesian,madigan1994model} where averaging the weights of multiple models fine-tuned with different hyperparameter configurations improves accuracy and robustness \cite{wortsman2022model}.
\begin{table}[H]
\begin{adjustbox}{width=\columnwidth, center}
\begin{tabular}{|l|}
\hline
climate change, 
climate,
fossil fuel,
fracking, 
energy,
oil,  
coal,  \\
mining, 
gas, 
carbon, 
power, 
footprint, 
solar, 
drilling, 
tri-city, \\
petroleum, 
renewable,  
global warming, 
emission,  
ecosystem,  \\
environment, 
greenhouse, 
ozone, 
radiation, 
bioenergy, \\
biomass, 
green energy, 
methane, 
pollution, 
forest,
planet, \\
earth,  
ocean, 
nuclear,  
ultraviolet, 
hydropower, 
hydrogen,\\
hydroelectricity, 
geothermal,
sustainable,
clean energy.
\\
\hline
\end{tabular}%
\end{adjustbox}
\caption{List of the keywords for data collection.} 
\label{tab:keywords}
\end{table}
\section{Data}
\label{dataset}
%\ti{Need to edit this section}
We collect $88,022$ climate related English ads focusing on the United States from January 2021 - January 2022 using Facebook Ad Library API\footnote{\url{https://www.facebook.com/ads/library/api}} with the keywords `climate change', `energy', `fracking', `coal'. 
To create the list of keywords for collecting ads about climate and oil \& gas industries, we read multiple articles about climate policy, environmental justice, climate change mentioning green/clean energy, transition from fossil fuel to renewable energy, coal dependent US states, protection of fossil-fuel workers and communities, and other climate debates, and made a list of repeating statements. Then, we consult two researchers in Computational Social Science and construct a list of relevant keywords. 
The full list of keywords is in Table \ref{tab:keywords}.
% The details of keyword selection procedure and the full list of keywords (Table \ref{tab:keywords}) are described in Appendix \ref{app:ds}. 
%
Our collected ads are written in English. For each ad, the API provides
the ad ID, title, ad description, ad body, funding entity, spend, impressions, distribution over impressions broken down by gender (male, female, unknown), age ($7$ groups), and location down to states in the USA. So far, we have $408$ unique funding entities whose stances are known based on their affiliation from their websites and Facebook pages. These funding entities are the source of supervision in our model. As we don't know the stance of the ads, we assign the same stance for all ads sponsored by the same funding entity. This way, we have $25,232$ ads whose stances are known. 
%We have duplicate content among those collected ads because the same ad has been targeted to different regions and demographics with unique ad id. Therefore, we have $5414$ unique ads.
\begin{figure}[htbp]
  \centering  
  \includegraphics[width= 1\columnwidth]{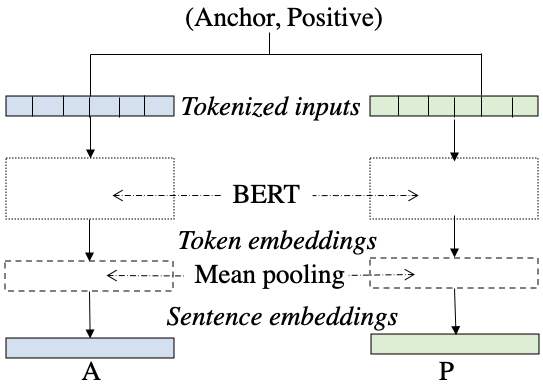}
\caption{Siamese-BERT network for contrastive learning to generate sentence embeddings.
}  
\label{fig:contrastive}
\end{figure}
\section{Problem Formulation}
We formulate our stance prediction problem as a minimally supervised model soup approach. We know the stance of the funding entity, but we don't know the stance of the ads. We assign the same stance for all ads sponsored by the same funding entity. 
% Let's assume, $X_a$ is the ads and $y_t$ is the assigned themes. 
We want to predict the stance of the ad using the model soup approach in the following way:
\begin{equation}\label{eq:1}
    \text{Point estimation: } P(y_s| X_a, \theta, y_t)
\end{equation}
% \begin{gather}\label{eq:2}
%     \text{Bayesian posterior: } \notag \\ 
%     P(\theta| y_s, X_a, y_t) \propto P(\theta) P(y_s| X_a, \theta, y_t)
% \end{gather}
\begin{equation}
  \begin{gathered}
    \text{Bayesian posterior: } \\ P(\theta| y_s, X_a, y_t) \propto P(\theta) P(y_s| X_a, \theta, y_t)
  \end{gathered}\label{eq:2}
\end{equation}
where, $X_a$ is the ad, $y_s$ is the predicted stance, $y_t$ is the assigned themes, $\theta$ is the model parameter. For the point estimation in Equation \ref{eq:1}, we fine-tuned the pre-trained BERT model \cite{devlin2019bert} by concatenating theme information. For Bayesian model averaging (Equation \ref{eq:2}), we implement both the uniform and greedy soup approaches provided by \citet{wortsman2022model} including messaging theme, which can be regarded as cheap Bayesian posterior approximations.
We get the theme $y_t$, using the contrastive learning approach following \citet{reimers2019sentence}.
% \section{Themes and Phrases Generation}
\section{Methodology}
In this section, we describe how to obtain sentence embedding using contrastive learning, generate themes and phrases, assign themes for the ad content, and implement model soup in our problem.

\subsection{Sentence Embeddings with Contrastive Learning}
\label{sec:finesbert}
We use $88k$ unlabeled ads for finetuning Sentence BERT (SBERT) \cite{reimers2019sentence}. Our training approach uses a siamese-BERT architecture during fine-tuning (Fig. \ref{fig:contrastive}).
% in Appendix \ref{app:fsbert}). 
During each step, we process a sentence $S$ (anchor) into BERT, followed by sentence $T$ (positive example). In our case, the anchor is the ad text, and a positive example is the ad description or ad summary. Some ads do not have ad descriptions. In that case, we generate an ad summary using BART summarizer \cite{lewis2020bart}. BERT generates token embeddings. Finally, those token embeddings are converted into averaged sentence embeddings using mean-pooling. Using the siamese approach, we produce two of these per step — one for the anchor $A$ and another for the positive called $P$. We use multiple negatives ranking loss which is a great loss function if we only have positive pairs, for example, only pairs of similar texts like pairs of paraphrases. In our case, positive pairs are ad text and description/summary. 

% We use $88k$ unlabeled ads for finetuning Sentence BERT (SBERT) \cite{reimers2019sentence}. Our training approach uses a siamese-BERT architecture during fine-tuning (Fig. \ref{fig:contrastive} in Appendix \ref{app:fsbert}). Please see Appendix \ref{app:fsbert} for the details. 
\begin{table}[t]
\begin{center}
 \scalebox{0.8}{\begin{tabular}{>{\arraybackslash}m{1.2cm}|>{\arraybackslash}m{9cm}}
 \toprule
 \textsc{\textbf{Pro-energy}} & Economy\_pro, Identity, Climate solution, Pragmatism, Patriotism, Against climate policy, Give away. \\
 \hline
 \textsc{\textbf{Clean-energy}} & Economy\_clean, Future generation, Environmental, Human health, Animals, Support climate policy, Alternative energy, Political affiliation.\\
 \bottomrule
\end{tabular}}
\caption{Resulting themes.}
\label{tab:reasons}
\end{center}
\end{table}

\begin{table*}
\centering
    \resizebox{\textwidth}{!}{%
    \begin{tabular}{|l|l|}
      \hline
      \textbf{Themes} & \textbf{Phrases} \\
      \hline
      \textbf{\textcolor{red}{Economy\_pro}} &  \makecell[l]{
      "Oil and gas will create more jobs",
    "Without oil and gas, there is no job", \\
    "Fracking supports thousands of jobs",
    "Without fracking, we will be jobless", \\
    "Oil and gas help local business",
    "Without oil and gas, our economy would be at risk", \\
    "Oil and gas industries pay high wages",
    "Jobs would be lower paid without oil and gas", \\
    "Local business would suffer without the oil and gas industry",
    "Don’t take jobs away from the coal miners", \\
    "Coal is powering economic progress", 
    "Protect our jobs", \\
    "Banning fossil fuels will lead to job losses",
    "Fracking jobs will bring new opportunities to rural areas", \\
    "Local communities would suffer due to the loss of tax revenue",
    "Natural gas ban would kill local jobs", \\
    "Oil and gas industries help the community through philanthropic efforts",
    "Energy industry gives back to communities", \\
    "Without the oil and gas industry, there would be less philanthropy". }    \\
      \hline
      \textbf{\textcolor{red}{Identity}} &  \makecell[l]{
      "Shifting away from fossil fuels is the loss of our culture",
   "Destruction of fossil fuel industry feels like the destruction of our identity", \\
   "Fossil fuel workers struggle with a loss of identity due to factory shut down",
   "We should protect our community identity", \\
   "Our identities are at stake",
   "Support the miners",
   "Coal is not just a Job, it’s a way of Life", \\
   "Remember the pride that coal mining gave us",
   "We are fighting for our identity", \\
   "Support our families and communities through supporting oil and gas industries".}    \\
      \hline
     \textbf{\textcolor{red}{ClimateSolution}} &  \makecell[l]{
     "We support reducing greenhouse gas emissions",
    "We develop technologies to reduce carbon emission", \\
    "We are committing to net-zero emissions",
    "We are transitioning energy mix away from fossil fuels", \\
    "We are moving towards renewables",
    "Natural gas is the future of clean energy", \\
    "Fossil gas is a low carbon energy source",
    "Natural gas is the perfect partner to renewables", \\
    "Natural gas is part of the solution to climate change",
    "Thanks to natural gas, emissions have reduced", \\
    "The oil and gas industry has to be a partner, not a problem",
    "Renewable natural gas will help us get to net zero carbon emissions as fast as we can".}    \\
      \hline
      \textbf{\textcolor{red}{Pragmatism}} &  \makecell[l]{
     "Oil and gas are affordable energy sources",
    "Without oil and gas, energy would be expensive", \\
    "Oil and gas are reliable energy sources",  
    "Oil and gas will keep the lights on no matter what", \\
    "Banning fossil fuel would make energy unreliable",
    "Without oil and gas, energy would be unreliable", \\
    "Oil and gas are safe",
    "Oil and gas power our lives",
    "Oil and gas are efficient", \\
    "Oil and gas meet our essential energy needs",
    "Oil and gas are resilient",
    "Oil and gas are abundant",
    "Oil and gas are secure".}    \\
     \hline
     \textbf{\textcolor{red}{Patriotism}} &  \makecell[l]{
    "Shutting down local oil and gas production would force us to increase reliance on unstable foreign oil", \\
    "We achieved record-high oil and gas production",
    "US is leading in oil and gas production", \\
     "US is an energy leader",
     "Without US oil and gas, the world would be forced to use dirtier emissions intensive oil and gas", \\
     "Stand up for American energy",
     "Keep Alaska competitive",
     "It's not patriotic to shut off American energy", \\
     "We don’t have to necessarily be reliant on the Middle East",
     "We are loaded with coal. It’s here and it’s ours".}    \\
     \hline
    \textbf{\textcolor{red}{AgainstClimatePolicy}} &  \makecell[l]{
    "The Build Back Better Act will ruin our economy",
    "Green New Deal would take America back to the dark ages", \\
    "Biden and Democrats own this energy crisis",
    "Biden's pipeline closure increases gas price", \\
    "Government's climate agenda is harmful to our economy",
    "Democrats' impractical energy policies won't stop climate change", \\
    "Government's climate policy is outrageous",
    "D.C. Socialists are attacking the oil and gas industry", \\
    "Biden's climate policy would make energy unaffordable".}    \\
     \hline
    \textbf{\textcolor{red}{GiveAway}} &  \makecell[l]{
    "We are giving away free gas",
    "Collect free coupon for gas".}    \\
     \hline
     \textbf{\textcolor{teal}{Economy\_clean}} &  \makecell[l]{
      "Compared with fossil fuel technologies, which are typically mechanized and capital intensive, the renewable energy industry is more labor intensive", \\
    "Fast-growing renewable energy jobs offer higher wages",
    "Fossil fuels are expensive", \\
    "Renewable energy opens up job opportunities",
    "Clean energy will create jobs boom", \\
    "Clean energy can rebuild our economy",
    "Nuclear energy can bring new clean energy jobs", \\
    "Losing nuclear power plants meaning losing jobs",
    "Make polluters pay to clean up their messes", \\
    "Energy companies put profit over people",
    "Big oil and gas companies are forcing American families to pay more".}    \\
      \hline
      \textbf{\textcolor{teal}{HumanHealth}} &  \makecell[l]{
      "Climate change is the single biggest health threat facing humanity", \\
    "Changing weather patterns are expanding diseases, and extreme weather events increase deaths and make it difficult for health care systems to keep up", \\
    "Our communities are facing increased risk of illness, disease, and even death from our changing climate", \\
    "Climate impacts are already harming health through air pollution, disease, extreme weather events, forced displacement, pressures on mental health, \\ and increased hunger and poor nutrition in places where people cannot grow or find sufficient food", 
    "Climate crisis is impacting our communities", \\
    "Fossil fuels threaten our health",
    "We need breathable air",
    "Toxic pollution kills people".}    \\
      \hline
     \textbf{\textcolor{teal}{FutureGeneration}} &  \makecell[l]{"Protect our children, family and future generations",
    "Climate change is a grave threat to children’s survival",
    "Clean air for healthier kids", \\
    "Children’s immune systems are still developing, leaving their rapidly growing bodies more sensitive to disease and pollution", 
    "Save the children",\\
    "Hotter temperatures, air pollution, and violent storms are leading to immediate, life-threatening dangers for children, \\ including difficulty breathing, malnutrition and higher risk of infectious diseases". 
     }    \\
      \hline
      \textbf{\textcolor{teal}{Environmental}} &  \makecell[l]{
     "Carbon dioxide and additional greenhouse gas emissions are leading contributors to climate change and global warming", \\
    "By slowing the effects of climate change and eventually reversing them, we can expect to see a reduction in extreme \\ weather like droughts, floods, and storms caused by global warming", "Protect our planet", \\
    "Changes in the climate and increases in extreme weather events are among the reasons behind a global rise in hunger and poor nutrition", \\
    "Changes in snow and ice cover in many Arctic regions have disrupted food supplies from herding, hunting, and fishing", \\
    "Destructive storms have become more intense and more frequent in many regions due to climate change", \\
    "Climate change is changing water availability, making it scarcer in more regions", \\
    "Global warming exacerbates water shortages in already water-stressed regions and is leading to an increased risk of \\
    agricultural droughts affecting crops, and ecological droughts increasing the vulnerability of ecosystems", \\
    "The rate at which the ocean is warming strongly increased over the past two decades, across all depths of the ocean", \\
    "Melting ice sheets cause sea levels to rise, threatening coastal and island communities", \\
    "More carbon dioxide makes the ocean more acidic, which endangers marine life and coral reefs", \\
    "As greenhouse gas concentrations rise, so does the global surface temperature", \\
    "Wildfires start more easily and spread more rapidly when conditions are hotter",
    "Protect our air",
    "Protect our ocean", \\
    "Climate crisis affects the environment",
    "The top cause contributing to carbon dioxide emissions is electricity generation from fossil fuel power plants".}    \\
     \hline
     \textbf{\textcolor{teal}{Animals}} &  \makecell[l]{
    "Climate change poses risks to the survival of species on land and in the ocean", \\
    "One million species are at risk of becoming extinct within the next few decades", \\
    "Toxic pollution kills animals",
    "Wildlife is severely affected by the reduction of rainfall and a lack of water", \\
    "In the U.S. and Canada, moose are struggling due to an increase in ticks and parasites that are surviving the shorter, milder winters".}    \\
     \hline
    \textbf{\textcolor{teal}{AltEnergy}} &  \makecell[l]{
    "Transitioning to renewable energy is not only necessary to fight the climate crisis, but also the only way we can quickly and effectively meet rising energy demands", \\
    "Alternative energy sources have a much lower carbon footprint than natural gas, coal, and other fossil fuels", \\
    "We can diversify our energy supply by implementing the widespread use of large-scale renewable energy technologies \\ and minimizing our imported fuel dependency",
    "Renewable energy is cheap",
    "Sustainable energy is the future".}    \\
     \hline
    \textbf{\textcolor{teal}{SupportClimatePolicy}} &  \makecell[l]{
    "The Build Back Better Act would put $\$555$ billion toward building a clean energy economy in the United States, \\
    the largest single investment in combating climate change in American history", "Support clean energy", \\
    "Green New Deal is a crucial framework for meeting the climate challenges we face",
    "Support the Energy Jobs \& Justice Act",
    "Stop corporate polluters", \\
    "Big oil and gas industries should be held accountable for climate change",
    "Join Regional Greenhouse Gas Initiative today",
    "Support climate policy", \\
    "Biden should honor his climate and justice commitments",
    "We need climate leader",
    "We need to hold our leaders accountable for climate crisis".}    \\
     \hline
     \textbf{\textcolor{teal}{PoliticalAffliation}} &  \makecell[l]{
    "Owners of oil and gas companies are the top donors to a political action committee", \\
    "Big oil and gas industries spend millions to fight climate bills".
    }    \\
     \hline
    \end{tabular}}
    % \caption{Pro-energy themes and phrases to show how the oil \& gas sector uses social media to influence the narrative on climate change. }
    \caption{Pro-energy (red) and clean-energy (green) themes and phrases to show how the sponsors use social media to influence the narrative on climate change.}
\label{tab:themesPro}
  %\end{center}
\end{table*}
\subsection{Themes and Phrases Generation}
To analyze climate campaigns, we model the climate related stance expressed in each ad (i.e., pro-energy, clean-energy) and the underlying reason behind such stance. For example, the top ad (brown box) of Fig.~\ref{fig:climate_ads} expresses a pro-energy stance and mentions their support for local jobs as the reason to take this stance. 

Three main challenges are involved in this analysis: 1) constructing the space of possible themes, 2) mapping ads to the relevant themes, and 3) predicting the stance leveraging the themes. 
We combine computational and qualitative techniques to uncover the most frequent themes cited for pro-energy and clean-energy  stances. We build on previous studies that characterized the arguments  supporting the oil and gas industries \cite{miller2016audience}. In this work, researchers develop four broad categories of pro-energy themes by looking at audience responses to ads from fossil fuel companies. 
As energy is an economic, social, security, and environmental concern, we go through relevant research conducted by \href{https://www.un.org/en/climatechange}{United Nations}, \href{https://influencemap.org/}{influencemap.org} and \href{https://www.pewresearch.org/science/2016/10/04/public-views-on-climate-change-and-climate-scientists/}{pewresearch.org} to construct a list of potential themes and phrases for each theme. We add new relevant pro-energy themes and corresponding phrases that were not covered by previous work, such as \textit{``Green New Deal would take America back to the dark ages"} which falls under a new theme called `\textbf{Against Climate Policy}'. As the initial set of themes contains mostly pro-energy arguments, we add reasons for supporting climate actions which are clean-energy themes, e.g., ``\textit{Climate change is a grave threat to children’s survival}" $\Rightarrow$ \textbf{Future Generation}. Then, we consult with two researchers in Computational Social Science and finalize the relevant themes with corresponding phrases. The final set of themes can be observed in Table \ref{tab:reasons}.
The full list of phrases for each theme can be observed in Table \ref{tab:themesPro}.
%at Appendix \ref{app:themes_phrases}.
% Table \ref{tab:themesPro} shows the full list of phrases for pro-energy themes. 

%Table \ref{tab:theme} shows the themes and the explanation of key narratives included in each theme.
% The full list of phrases for each theme can be observed in Appendix \ref{app:themes_phrases}.
% Table \ref{tab:themesPro} shows the full list of phrases for pro-energy themes. 
\subsection{Assign Themes}
Our main goal is to ground these themes in a set of approximately $25k$ labeled (stance) ads. To map ads to themes, we use the cosine similarity between their fine-tuned sentence BERT embeddings (details of fine-tuning provided in subsection \ref{sec:finesbert}) of the ad text and the phrases of each theme. To check the quality of the theme label, we annotated around $300$ ads with corresponding themes and noticed an accuracy of $38.4\%$ and macro-avg F1 score of $40.2\%$, which is better than the random ($6.6\%$).
\subsection{Bayesian Model Averaging}
In this work, we develop a minimally supervised model soup approach by incorporating messaging themes to identify the stances of climate ads on Facebook. We used two approaches for model soup. The first one is uniform soup \cite{wortsman2022model}. We consider a neural network $f(x,\theta)$ with input data $x$ and parameters $\theta$. For uniform soup, we take the average of the fine-tuned model parameters ($f(x, \frac{1}{k} \sum_{i=1}^k\theta_i)$) where $\theta_i$ can be considered as samples from the Bayesian posterior and the average can be viewed as a cheap approximation to Bayesian model average. The second one is the greedy soup approach \cite{wortsman2022model}. For the greedy soup, we first sort the models in decreasing order of validation set accuracy. The soup is constructed by sequentially adding each model as a potential ingredient in the soup and only keeping the model in the soup if performance on the validation set improves.
\begin{table}
\centering
\resizebox{.75\columnwidth}{!}{%
\begin{tabular}{lcc}
    \toprule
      \textbf{Data split} & \textbf{Number of Funding entities}  & \textbf{Number of Ads} \\
     \midrule
      Training & 261 & 17780 \\
      Validation & 65 & 2074  \\
      Testing & 82 & 5378 \\
    \bottomrule
    \end{tabular}}%
\caption{Data details. }
\label{tab:data}
\end{table}
\begin{table}
\centering
\resizebox{1\columnwidth}{!}{%
\begin{tabular}{llccc}
    \toprule
      \textbf{Model} & \textbf{Method} & \textbf{Accuracy}  & \textbf{Macro-avg F1} \\
     \midrule 
     % RoBERTa (text) & 0.900 & 0.802  \\
     % BERT (text) & 0.909 & 0.866  \\
     % \textit{Uniform Model soup (text)} & \textit{0.943} & \textit{0.880} \\
     % \textit{Greedy Model soup (text)} &  \textit{0.933} & \textit{0.872} \\
      LR\_tf-idf & Best individual model & 0.810 & 0.506  \\
      \hline
      RoBERTa-base & Best individual model & 0.943 & 0.879  \\
      \hline
      T5-small & Best individual model &  0.874 & 0.8743 \\
      \hline
      BERT-base & Best individual model & 0.921 & 0.854 \\
      & \textit{\textbf{Uniform Model soup}} &  0.944 & \textbf{0.888 } \\
      & \textit{\textbf{Greedy Model soup}} & \textbf{ 0.945} & 0.884 \\
    \bottomrule
    \end{tabular}}%
% \caption{Model performance for stance prediction using minimally supervised model soup approach. FBERT: Fine-tuned of pre-trained BERT model, FE: Funding entity, Hyper: Hyperparameter. }
\caption{ Performance comparison on test data. Comparing model soup with simple Logistic Regression with tf-idf feature (LR\_tf-idf) as well as standalone BERT, RoBERTa, and T5 baselines.}
\label{tab:result}
\end{table}
\begin{table}
\centering
\resizebox{1\columnwidth}{!}{%
\begin{tabular}{lcccc}
    \toprule
      \textbf{Model} & \textbf{Accuracy}  & \textbf{Macro-avg F1} & \textbf{Learning rate} &  \textbf{Weight decay}\\
     \midrule
      FBERT\_Hyper1 (text) & 0.897 & 0.833 & 2.00E-05 & 0.01 \\
      FBERT\_Hyper2 (text) & 0.909 & 0.866 & 1.00E-05 & 0.01 \\
      FBERT\_Hyper3 (text) & 0.899 & 0.687 & 1.00E-04 & 0.001 \\
      FBERT\_Hyper4 (text) & 0.895 & 0.774 & 1.00E-04 & 0.01 \\
      FBERT\_Hyper5 (text) & 0.905 & 0.856 & 1.00E-05 & 0.001 \\
      FBERT\_Hyper6 (text) & 0.898 & 0.813 & 3.00E-05 & 0.001 \\
      FBERT\_Hyper7 (text) & 0.896 & 0.825 & 3.00E-05 & 0.01 \\
      FBERT\_Hyper8 (text) & 0.892 & 0.833 & 2.00E-05 & 0.1 \\
      FBERT\_Hyper9 (text) & 0.885 & 0.813 & 1.00E-04 & 0.0001 \\
      FBERT\_Hyper10 (text) & 0.906 & 0.861 & 1.00E-05 & 0.1 \\
     \textit{Uniform Model soup (text)} & \textit{0.943} & \textit{0.880} & - &  - \\
    \textit{Greedy Model soup (text)} &  \textit{0.933} & \textit{0.872} & - &  - \\
      Point\_est\_Hyper1 (text + thm)  & 0.921 & 0.854 & 2.00E-05 & 0.01 \\
      Point\_est\_Hyper2 (text + thm) & 0.883 & 0.835 & 1.00E-05 & 0.01 \\
      Point\_est\_Hyper3 (text + thm) & 0.916 & 0.695 & 1.00E-04 & 0.001 \\
      Point\_est\_Hyper4 (text + thm) & 0.874 & 0.845 & 1.00E-04 & 0.01 \\
      Point\_est\_Hyper5 (text + thm) & 0.897 & 0.826 & 1.00E-05 & 0.001 \\
      Point\_est\_Hyper6 (text + thm) & 0.902 & 0.825 & 3.00E-05 & 0.001 \\
      Point\_est\_Hyper7 (text + thm) & 0.894 & 0.830 & 3.00E-05 & 0.01 \\
      Point\_est\_Hyper8 (text + thm) & 0.894 & 0.829 & 2.00E-05 & 0.1 \\
      Point\_est\_Hyper9 (text + thm) & 0.888 & 0.781 & 1.00E-04 & 0.0001 \\
      Point\_est\_Hyper10 (text + thm) & 0.879 & 0.822 & 1.00E-05 & 0.1 \\
      \textit{\textbf{Uniform Model soup (text + thm)}} &  0.944 & \textbf{0.888 }& - &  -  \\
      \textit{\textbf{Greedy Model soup (text + thm)}} & \textbf{ 0.945} & 0.884 & - &  -  \\
    \bottomrule
    \end{tabular}}%
% \caption{Model performance on test data using minimally supervised model soup approach. FBERT: Fine-tuned of pre-trained BERT model, Point\_est: Point estimation, thm: Theme, Hyper: Hyperparameter. }
\caption{Ablation study. FBERT: Fine-tuned pre-trained BERT model, Point\_est: Point estimation, thm: Theme, Hyper: Hyperparameter. }
\label{tab:ablation}
\end{table}
% \section{Results and Analysis}
% \section{Experimental Details and Analysis}
\section{Experimental Details}
This section presents the experimental details of the stance prediction task on climate change-related ads. We randomly split our data based on the funding entity so that the same ads do not appear in the other splits. 
% Details number of funding entities and ads for each split are shown in Appendix \ref{app:exp} (Table \ref{tab:data}).
At first, we randomly split $20\%$ of the funding entities and keep them as a testing set. Then we randomly split the rest of the data and keep $20\%$ of that as a validation set and the rest as the training set. Details number of funding entities and ads for each split are shown in Table \ref{tab:data}.
We fine-tune the pre-trained BERT-base-uncased model \cite{devlin2019bert} and run for $10$ epochs for each hyperparameter setting, i.e., learning rate and weight decay. We set the maximum text sequence
length to $110$, batch size $32$, and use Adam optimizer \cite{kingma2015adam}. We concatenate the assigned theme with ad text so that our model can leverage the theme information. 
% Experimental details are shown in Appendix \ref{app:setup}.

We use pre-trained weights from the Huggingface Transformers library \cite{wolf2020transformers}. Evaluation is conducted once at the end of the training, without early stopping. We use a single GPU GeForce GTX 1080 Ti GPU, with 6 Intel Core i5-8400 CPU @ 2.80 GHz processors to run each model, and it takes around 15 minutes to run each model. But averaging several of these models to form a model soup requires no additional training and adds no cost at inference time. 
\begin{figure*}[htbp]
\begin{subfigure}{1\textwidth}
  \centering
  \includegraphics[width=\textwidth]{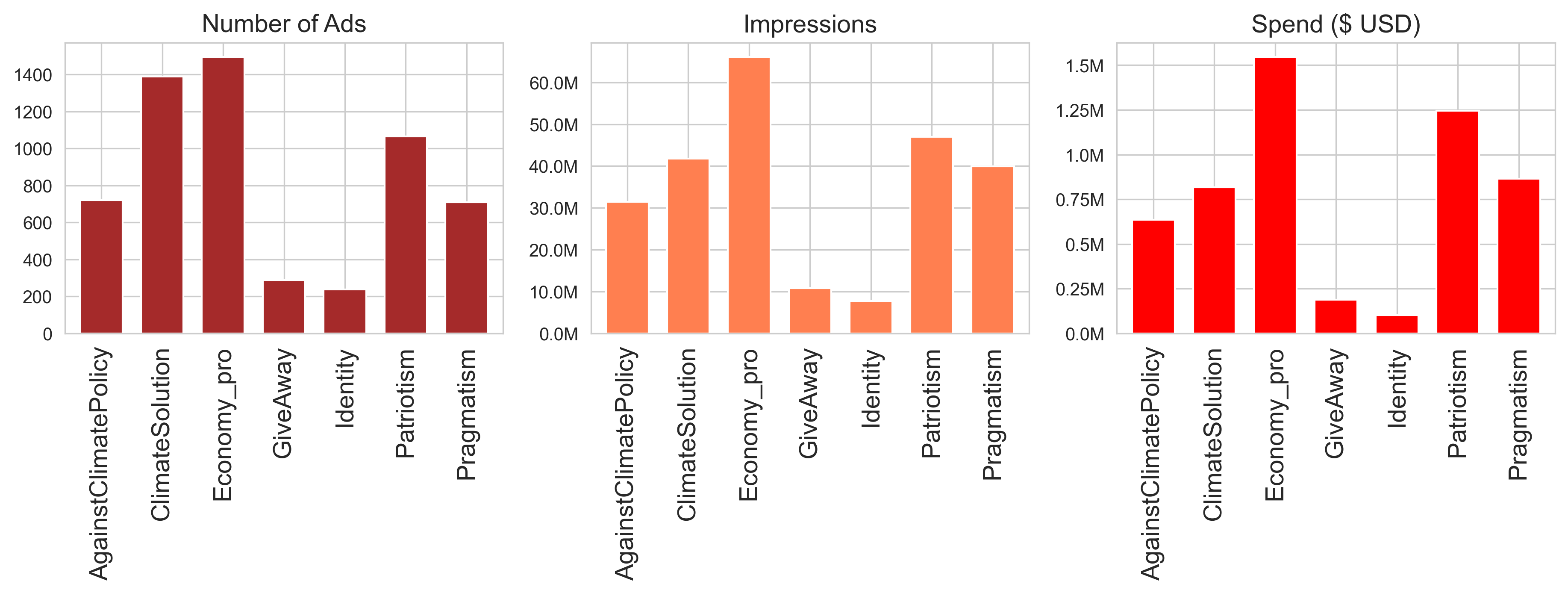}
  \caption{Pro-energy ads}
  \label{fig:ais_pro}
\end{subfigure}
\begin{subfigure}{1\textwidth}
  \centering
  \includegraphics[width=\textwidth]{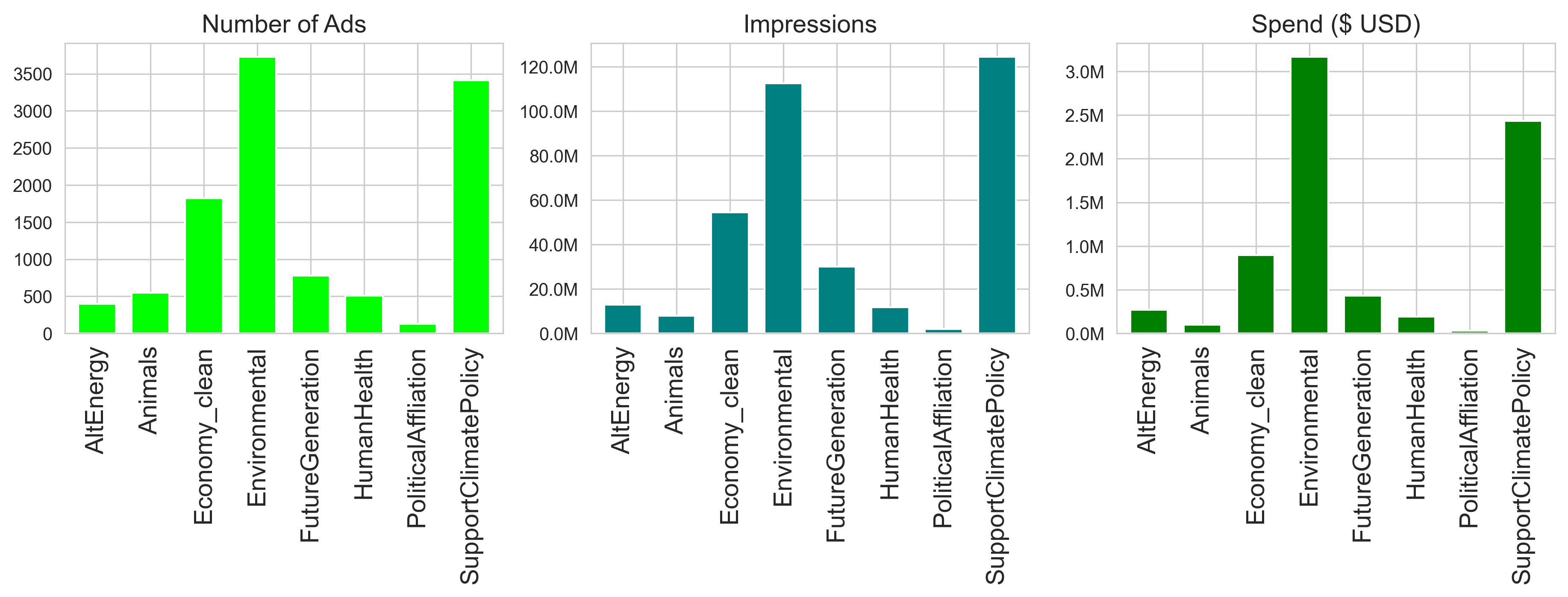}
  \caption{Clean-energy ads}
  \label{fig:ais_cln}
\end{subfigure}
\caption{Distribution of ad themes by Number of Ads, Impressions, and Spend.}
\label{fig:ais}
\end{figure*}
\begin{figure*}[htbp]
\begin{subfigure}{.35\textwidth}
  \centering
  \includegraphics[width=\textwidth]{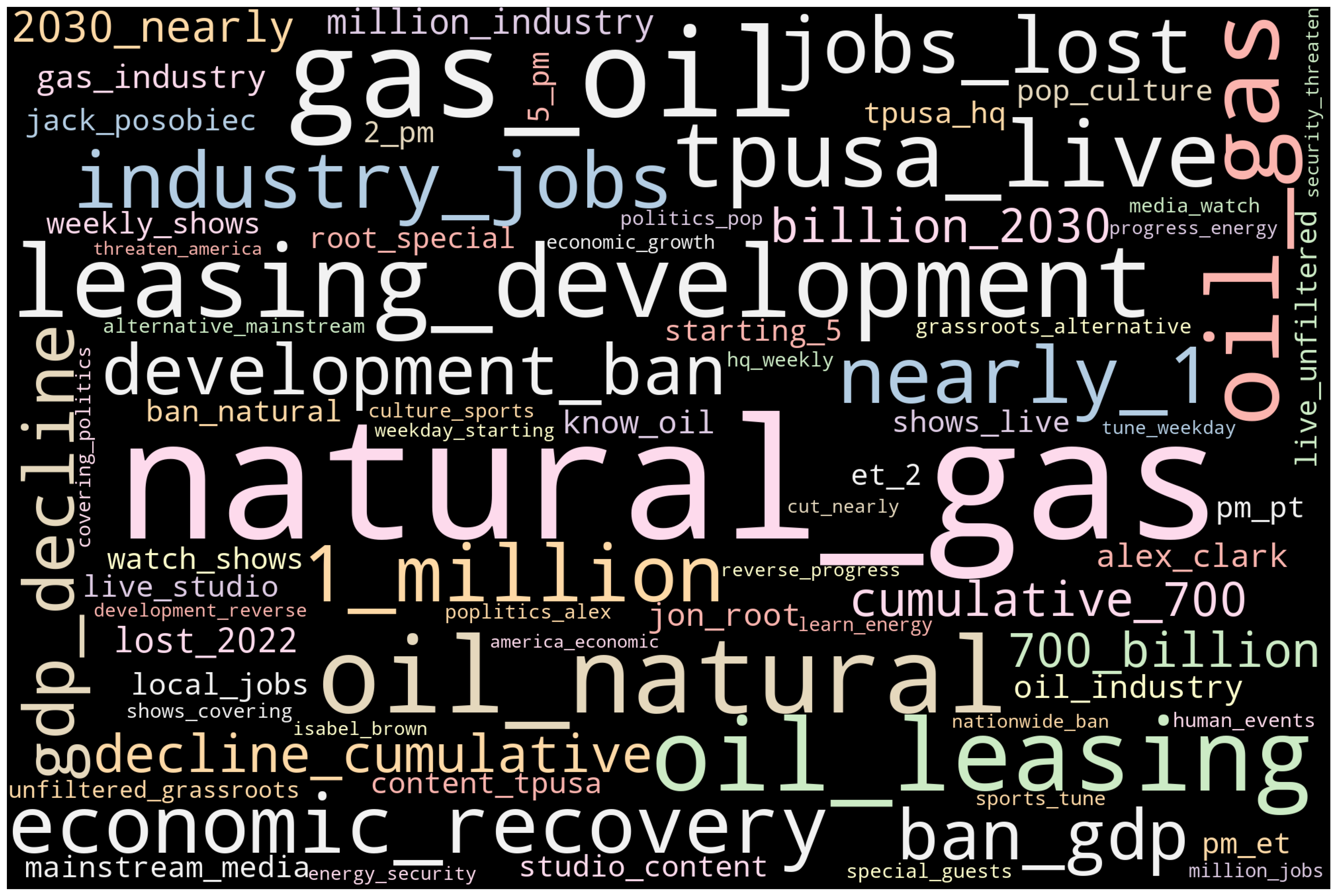}
  \caption{Economy\_pro}
  \label{fig:eco_pro}
\end{subfigure}%
\begin{subfigure}{.35\textwidth}
  \centering
  \includegraphics[width=\textwidth]{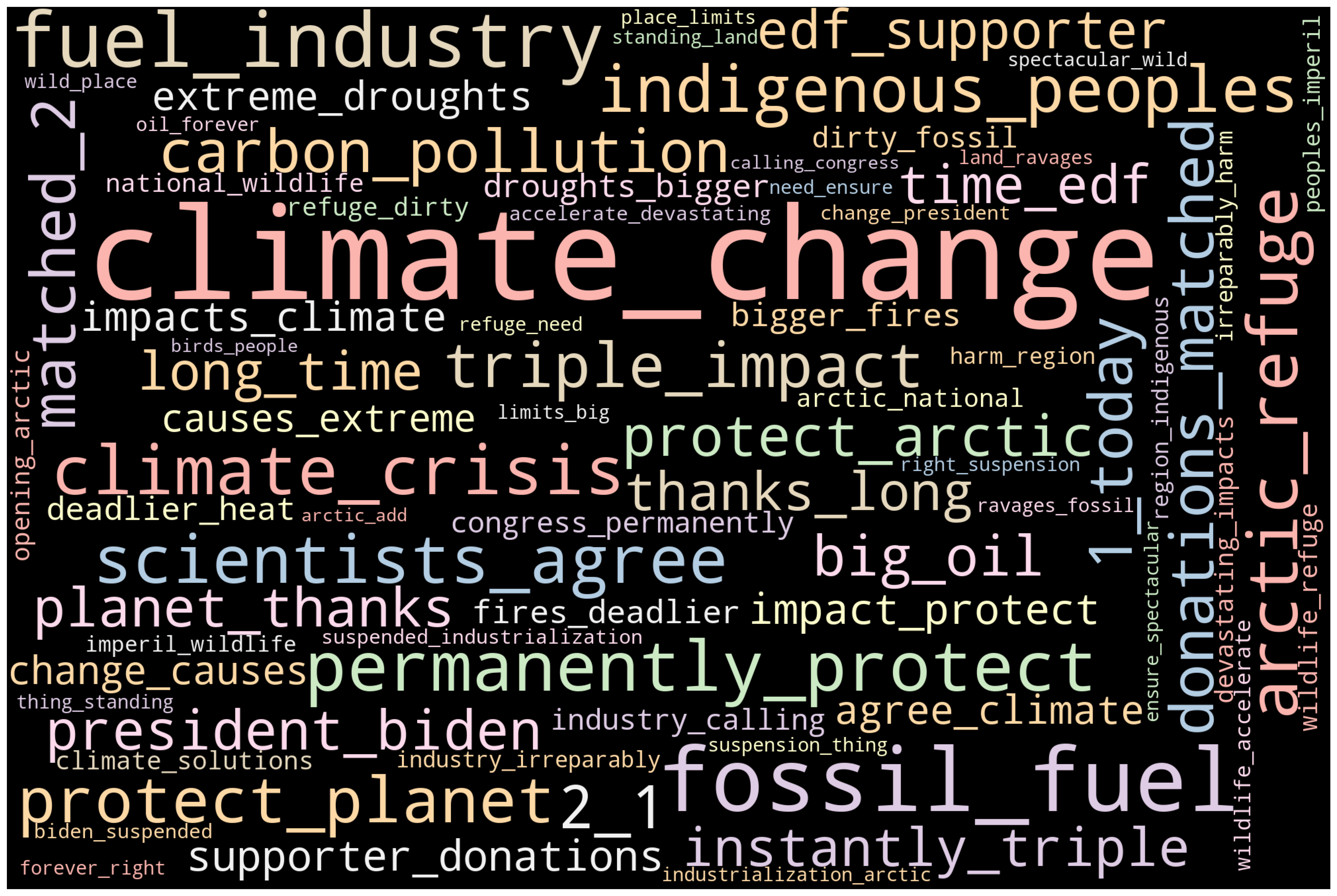}
  \caption{Environmental}
  \label{fig:env}
\end{subfigure}%
\begin{subfigure}{.35\textwidth}
  \centering
  \includegraphics[width=\textwidth]{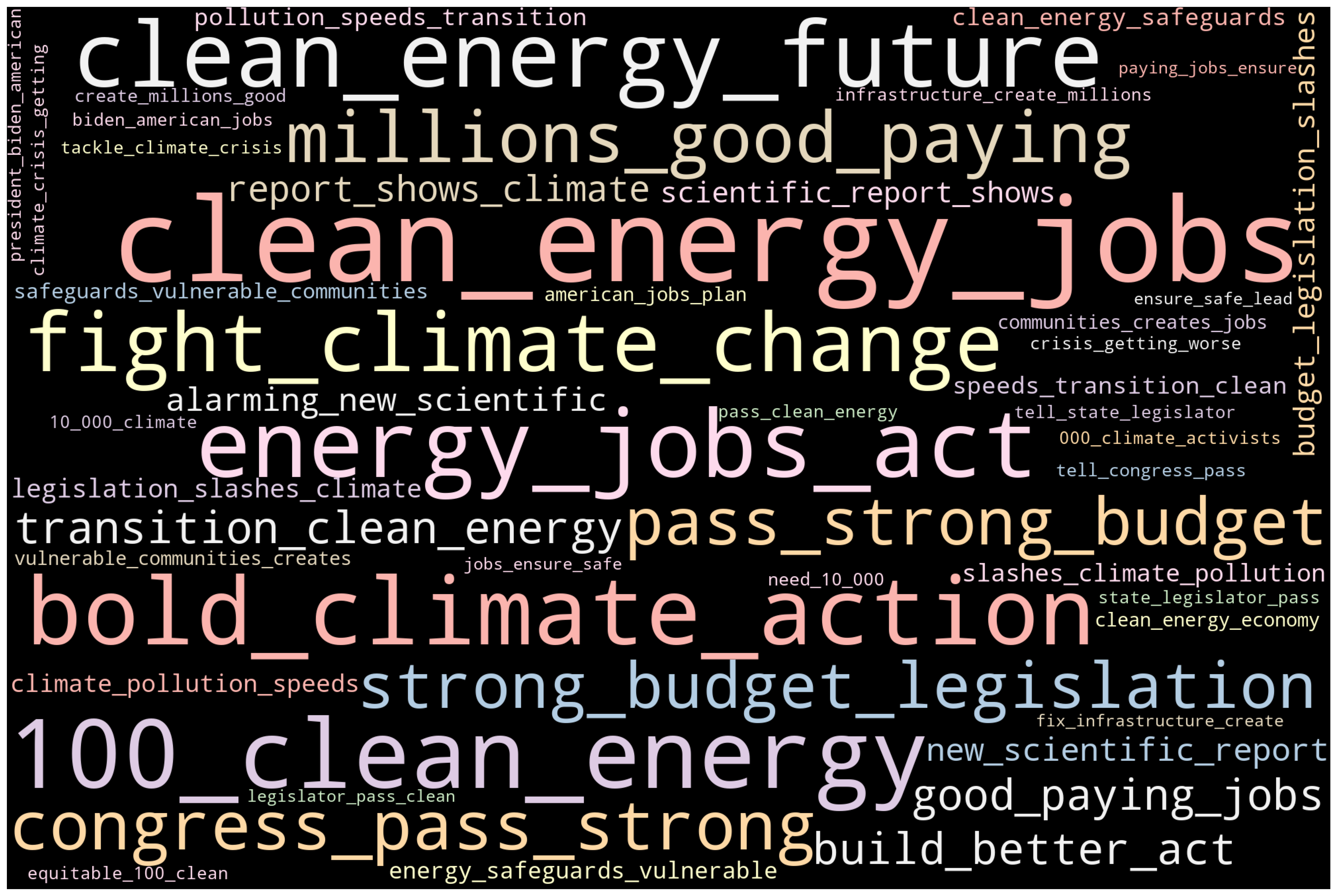}
  \caption{SupportClimatePolicy}
  \label{fig:sup_policy}
\end{subfigure}
\caption{Wordcloud for three messaging themes based on the popularity of ad impressions, expenditure, and the number of sponsored ads for both pro-energy and clean-energy ads.}
\label{fig:wc}
\end{figure*}
\subsection{Results}
We provide experimental results in Table \ref{tab:result}. For the evaluation metrics, we use accuracy and macro-average F1 score. 
% We notice that the uniform model soup using ad text + theme ($88.8\%$ macro-avg F1 score) outperforms the greedy model soup for text + theme and the best single models (Table \ref{tab:result}) (Answer to \textbf{RQ1}). 
At first, we compare our approach with simple Logistic Regression (LR) \cite{cox1958regression} trained on term frequency–inverse document frequency (tf-idf) features baseline (Table \ref{tab:result}). Then, to make sure that the model soup being a better hypothesis holds irrespective of the underlying language model (LM) architecture, we test our work on larger pre-trained LM, i.e., RoBERTa \cite{liu2019roberta}, T5 \cite{raffel2020exploring} besides BERT. Finally, we compare the performance accuracy and macro-average F1 score with the standalone models (best individual model) with respect to the model soup (Table \ref{tab:result}). From Table \ref{tab:result}, we notice that the uniform model soup using ad text + theme ($88.8\%$ macro-avg F1 score) outperforms the greedy model soup for text + theme and the best individual model baselines (Answer to \textbf{RQ1}). 

% while the ablation shows that a model soup outperforms the standalone �netuned pretrained BERT, it would also be ideal to test this on a larger/newer pretrained LM and compare the performance accuracy with the standalone model there w.r.t. model soup. Essentially, making sure that the model soup being better hypothesis holds irrespective of the underlying LM architecture. Might also be interesting to see how the few shot models take up this task instead of the model soup approach.

\begin{figure*}[htbp]
\begin{subfigure}{.5\textwidth}
  \centering
  \includegraphics[width=\textwidth]{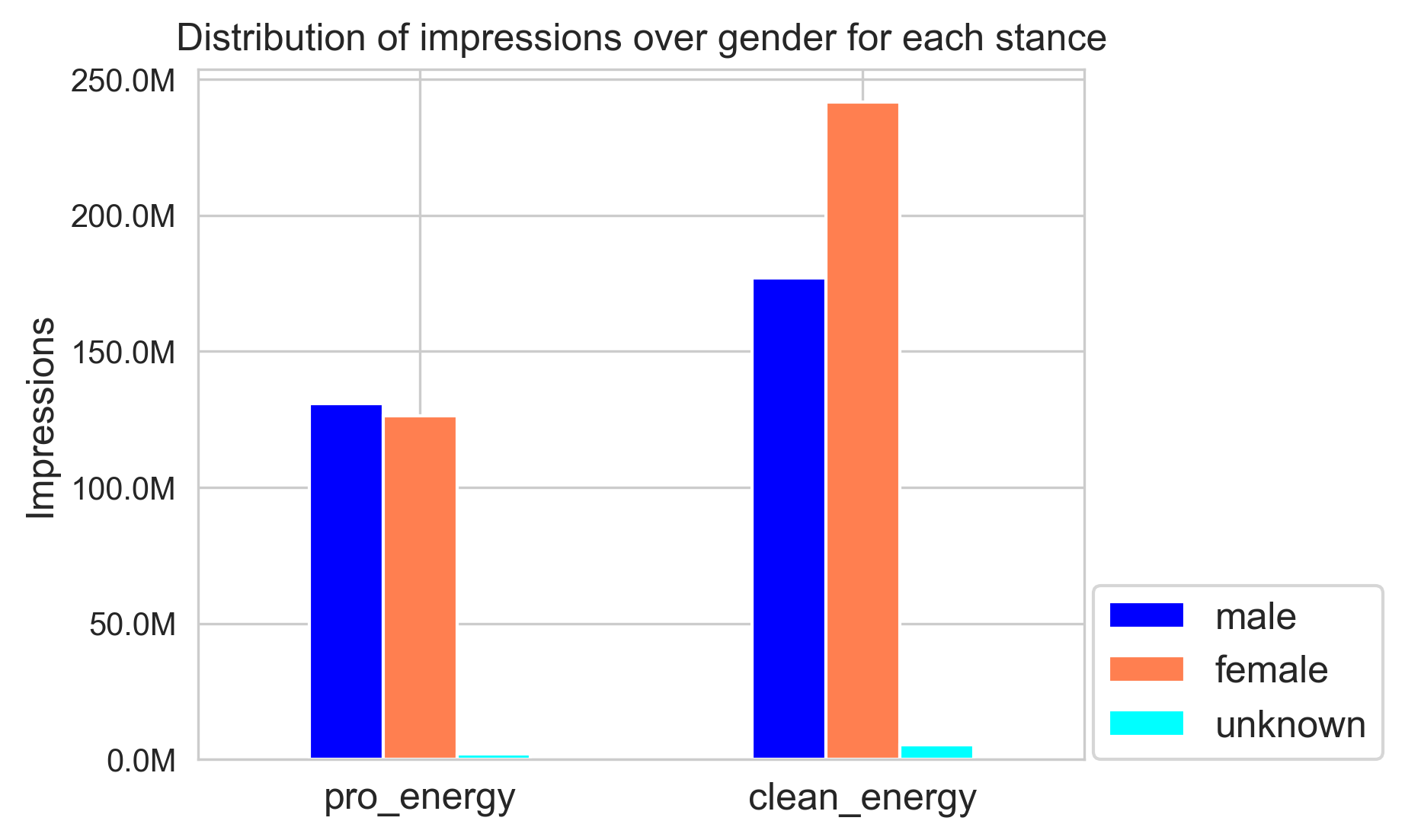}
  \caption{Gender}
  \label{fig:sgi}
\end{subfigure}%
\begin{subfigure}{0.5\textwidth}
  \centering
  \includegraphics[width=\textwidth]{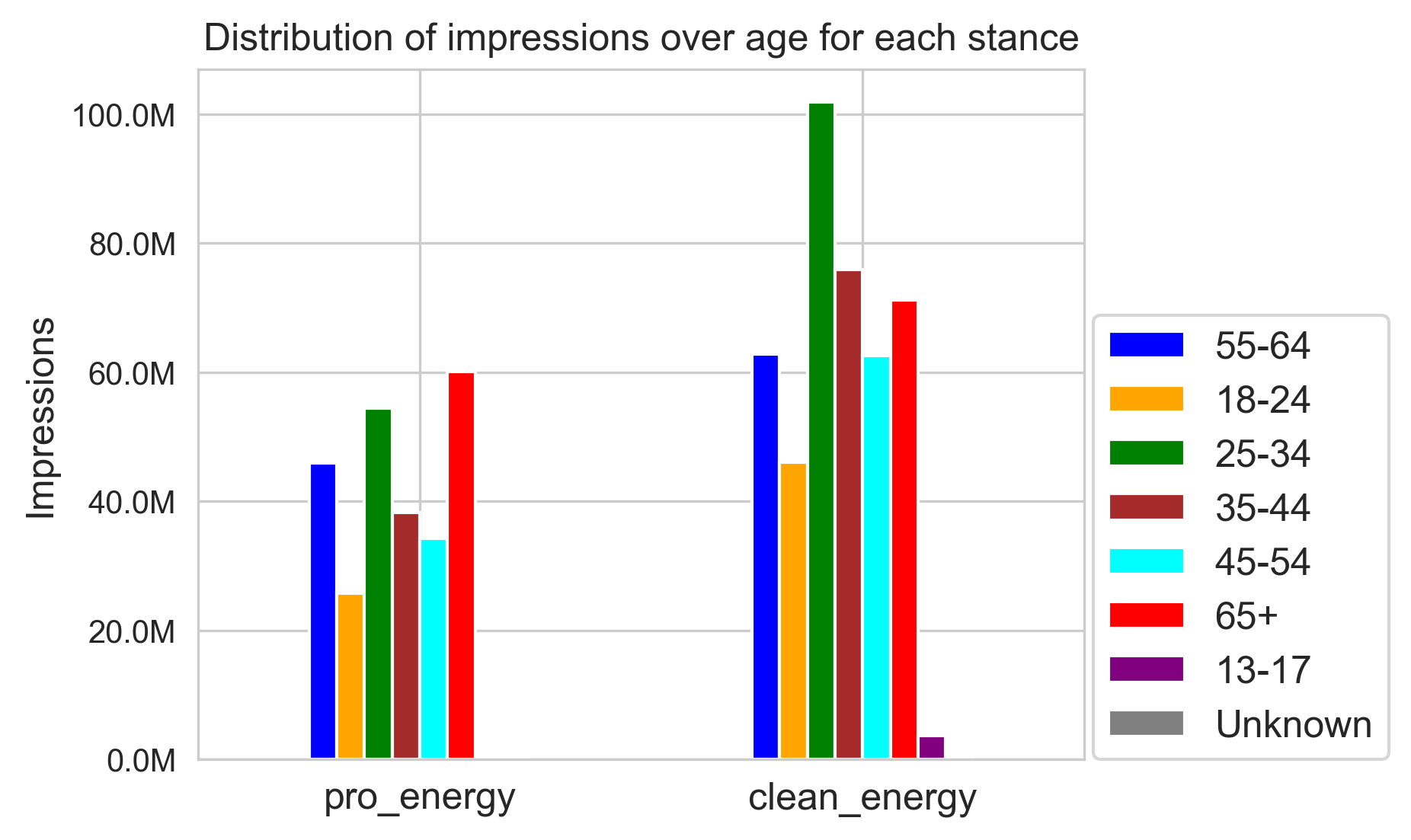}
  \caption{Age group}
  \label{fig:sai}
\end{subfigure}
\caption{Distribution of impressions over demographic distribution both for pro-energy and clean-energy ads. (a) More males than females watch the pro-energy ads. On the other hand, more females than males view clean-energy ads. (b) The older population ($65+$) watches the pro-energy ads. In contrast, the younger population ($25-34$) watches clean-energy ads.}
\label{fig:sdi}
\end{figure*}
\begin{figure*}[htbp]
\begin{subfigure}{.5\textwidth}
  \centering
  \includegraphics[width=\textwidth]{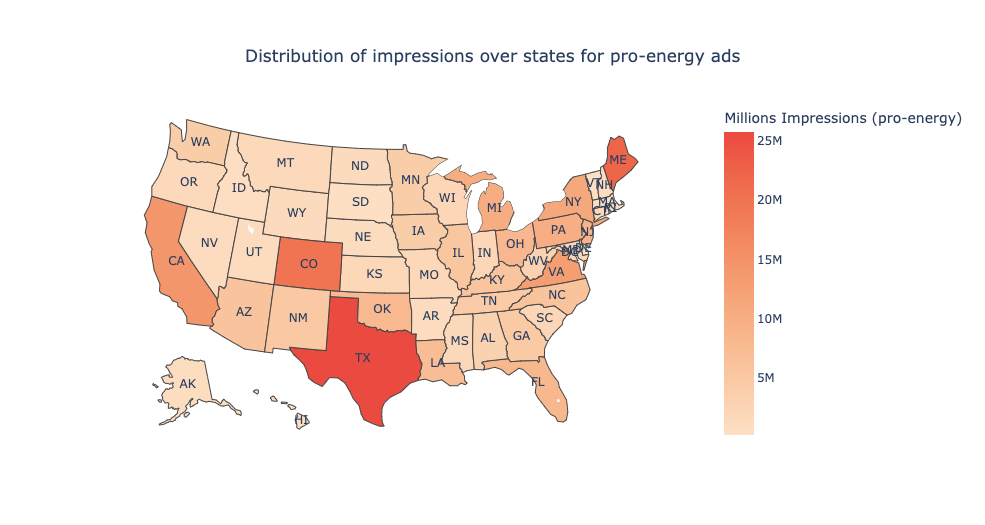}
  \caption{pro-energy}
  \label{fig:map_pro}
\end{subfigure}%
\begin{subfigure}{0.5\textwidth}
  \centering
  \includegraphics[width=\textwidth]{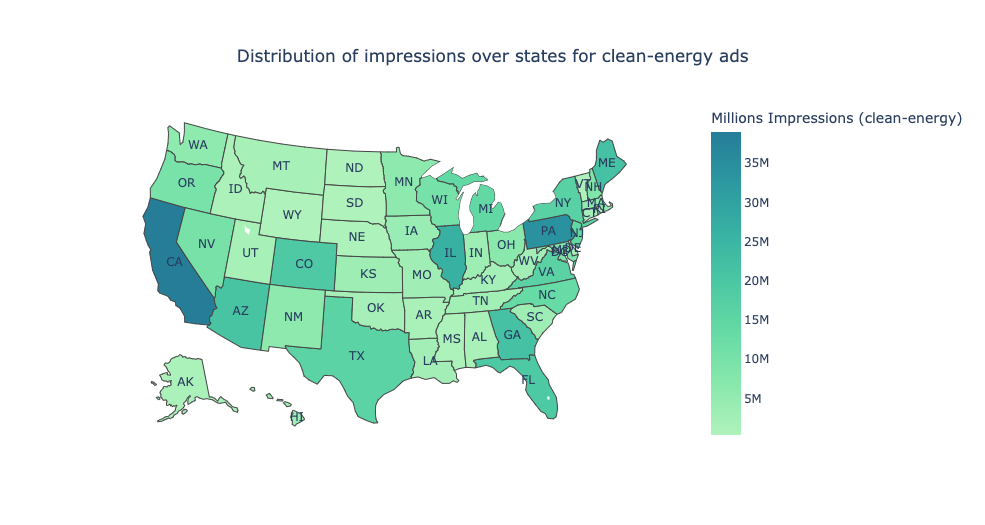}
  \caption{clean-energy}
  \label{fig:map_cln}
\end{subfigure}
\caption{Distribution of impressions over geographic. Pro-energy ads are mostly viewed from Texas (a), whereas clean-energy ads are mostly viewed from California (b).}
\label{fig:map}
\end{figure*}
\subsection{Ablation Study}
For the ablation study, we run the experiments using only ad text (we \textbf{do not} provide any theme information). We notice that the uniform model soup (text + theme) still gives better performance than the uniform model soup (text), greedy model soup (text), and the best single text only models (Table \ref{tab:ablation}).

\section{Analyses}
In this section, we present analyses that address our three research questions (\textbf{RQ2}, \textbf{RQ3}, and \textbf{RQ4}).

In subsection \ref{na}, we find that various advertisers prioritize distinct themes to promote their narratives that endorse particular stances. In subsection \ref{demo}, we find that advertisers aim their messages at particular demographics and geographic locations to spread their viewpoints. Subsection \ref{fe_type} shows that how messaging differs based on the entity type.

\subsection{Narrative Analysis}
\label{na}
We consider only ads with correct stance prediction and corresponding themes for narrative analysis.
To answer \textbf{RQ2}, we analyze the messaging strategies used by the advertisers (Fig. \ref{fig:ais}). By impressions and expenditures, the most popular \textbf{\textit{pro-energy}} messaging theme is `\textbf{Economy\_pro}', accounting for approximately $27\%$ of total impressions and $28.7\%$ of total expenditure (Fig. \ref{fig:ais_pro}). Under this theme, narratives promote how \textit{`natural gas and oil industry will drive economic recovery', `GDP would decline by a cumulative $700$ billion through 2030 and $1$ million industry jobs would be lost by 2022 under natural gas and oil leasing and development ban'} (Fig. \ref{fig:eco_pro}).

Based on impression, the most popular \textbf{\textit{clean-energy}} messaging category is `\textbf{SupportClimatePolicy}' (Fig. \ref{fig:ais_cln}) (approximately $35\%$), which features narratives supporting Build Back Better Act\footnote{\url{https://www.whitehouse.gov/build-back-better/}} to \textit{fight climate change, create clean energy jobs, equitable clean energy future, take bold climate action} (Fig. \ref{fig:sup_policy}). Based on spend, the most popular ($42\%$) \textbf{\textit{clean-energy}} messaging theme is `\textbf{Environmental}' (Fig. \ref{fig:ais_cln}). This theme focuses on narratives about \textit{`how dirty fossil fuel industries would harm the indigenous peoples and wildlife', `why climate scientists agree that climate change causes more extreme droughts, bigger fires and deadlier heat', `effects of carbon pollution on climate crisis'} etc (Fig. \ref{fig:env}). 
% \begin{figure*}[htbp]
%   \centering  
%   \includegraphics[width= 1\textwidth]{thm_fe.png}
% \caption{Pro-energy ad themes by funding entity type.}
%     \label{fig:thm_fe}
% \end{figure*}
% \begin{figure*}[htbp]
%   \centering  
%   \includegraphics[width= 1\textwidth]{ad_spend_pro_cln.png}
% \caption{Top 5 funding entities based on expenditure. Orange plot represents pro-energy. Green plot represents clean-energy. }
%     \label{fig:cost}
% \end{figure*}
\begin{table}[]
    \centering
    \resizebox{1\columnwidth}{!}{%
    \begin{tabular}{|l|l|}
      \hline
      \thead{\textbf{Type}} & \thead{\textbf{Entity}} \\
      \hline
      \textbf{Corporation} & EXXON MOBIL CORPORATION \\
      \textbf{Corporation} & Shell \\
      \textbf{Corporation} & BP CORPORATION NORTH AMERICA INC. \\
      \textbf{Corporation} & Twin Metals Minnesota \\
      \textbf{Corporation} & Wink to Webster Pipeline LLC \\
      \textbf{Industry Association} & AMERICAN PETROLEUM INSTITUTE \\
      \textbf{Industry Association} & New York Propane Gas Association \\
      \textbf{Industry Association} & Texas Oil \& Gas Association \\
      \textbf{Industry Association} & New Mexico Oil and Gas Association \\
      \textbf{Industry Association} & National Propane Gas Association \\
      \textbf{Advocacy Group} & Coloradans for Responsible Energy Development \\
      \textbf{Advocacy Group} & Grow Louisiana Coalition \\
      \textbf{Advocacy Group} & Voices for Cooperative Power \\
      \textbf{Advocacy Group} & Consumer Energy Alliance \\
      \textbf{Advocacy Group} & Maine Affordable Energy \\
    \hline
    \end{tabular}}
    \caption{List of entities from pro-energy ads.}
    \label{tab:entity}
\end{table}
\subsection{Demographic and Geographics Distribution by Impressions}
\label{demo}
As Facebook enables its customers to target ads using demographics and geographic information, we further analyze the distribution of the messaging categories to answer \textbf{RQ3}. 
At first, we perform a chi-square test \cite{cochran1952chi2} of contingency to calculate the statistical significance of an association between demographic group and their stances. The null hypothesis $H_0$ assumes that there is no association between the variables, while the alternative hypothesis $H_a$ claims that some association does exist. The chi-square test statistic is computed as follows:
\begin{align*}
    \chi^2 = \sum \frac{(observed - expected)^2 } {expected} 
\end{align*}
The distribution of the statistic $\chi^2$ is denoted as $\chi^{2}_{(df)}$, where $df$ is the number of degrees of freedom. $df = (r-1)(c-1)$, where $r$ represents the number of rows and $c$ represents the number of columns in the contingency table. The p-value for the chi-square test is the probability of observing a value at least as extreme as the test statistic for a chi-square distribution with $(r-1)(c-1)$ degrees of freedom.
To perform a chi-square test, we take gender distribution over stance and age distribution over stance separately to build contingency tables correspondingly.
\newline
The null hypothesis, $H_0$: whether the demographic group and their stances are independent, i.e., \textit{no relationship}.
\newline
The alternative hypothesis $H_a$: whether the demographic group and their stances are dependent, i.e., \textit{$\exists$ a relationship}.
\newline
We choose the value of significance level, $\alpha = 0.05$. The p-value for both cases is $< 0.05$, which is statistically significant. We reject the null hypothesis $H_0$, indicating some association between the audience's demographics and their stances on climate change. 
Fig. \ref{fig:sgi} shows that \textit{more males than females} view the \textbf{\textit{pro-energy}} ads, and \textit{more females than males} watch \textit{\textbf{clean-energy}} ads.
However, \textit{\textbf{pro-energy}} ads are mostly viewed by the \textit{older population ($65+$)} (Fig. \ref{fig:sai}). On the other hand, \textit{young people} from the age range of \textit{$25-34$} watch \textit{\textbf{clean-energy}} ads (Fig. \ref{fig:sai}).

In Fig. \ref{fig:map}, we show the distribution of impressions over US states for both stances. To plot the distribution, we use the Choropleth map\footnote{\url{https://plotly.com/python/choropleth-maps/}} in Python. \textbf{\textit{Pro-energy}} ads receive the most views from Texas which is the energy capital of the world\footnote{\href{https://www.eia.gov/todayinenergy/detail.php?id=49356}{www.eia.gov/}} (Fig. \ref{fig:map_pro}). Fig. \ref{fig:map_cln} shows that \textbf{\textit{clean-energy}} ads are mostly viewed from California because recently, CA has become one of the loudest voices in the fight against climate change\footnote{\href{https://www.pewtrusts.org/en/research-and-analysis/blogs/stateline/2022/10/06/california-takes-leading-edge-on-climate-laws-others-could-follow}{www.pewtrusts.org}}.
\begin{figure*}[htbp]
  \centering  
  \includegraphics[width= 1\textwidth]{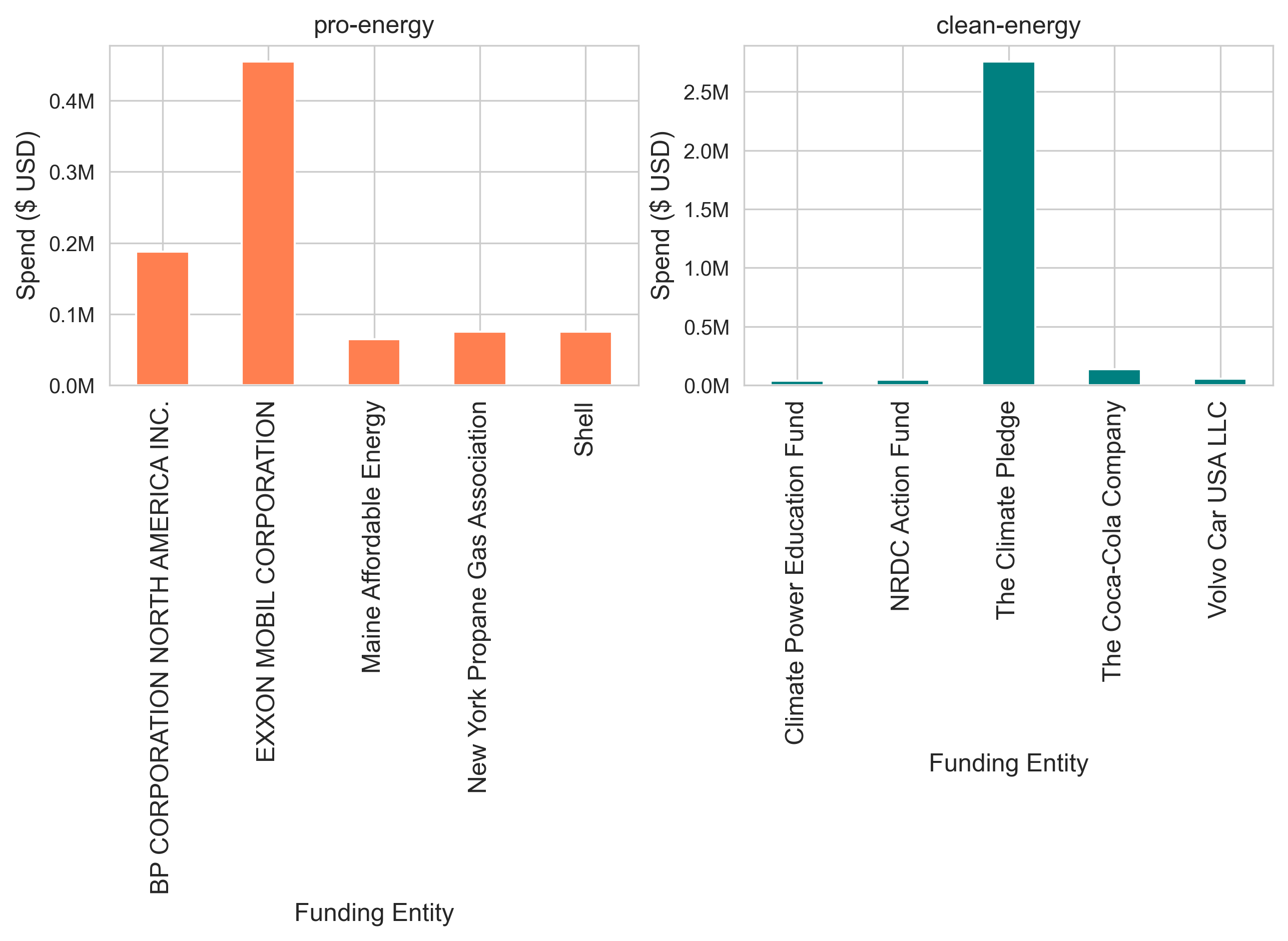}
\caption{Top 5 funding entities based on expenditure. Orange plot represents pro-energy. Green plot represents clean-energy. }
    \label{fig:cost}
\end{figure*}
\begin{figure*}[htbp]
  \centering  
  \includegraphics[width= 1\textwidth]{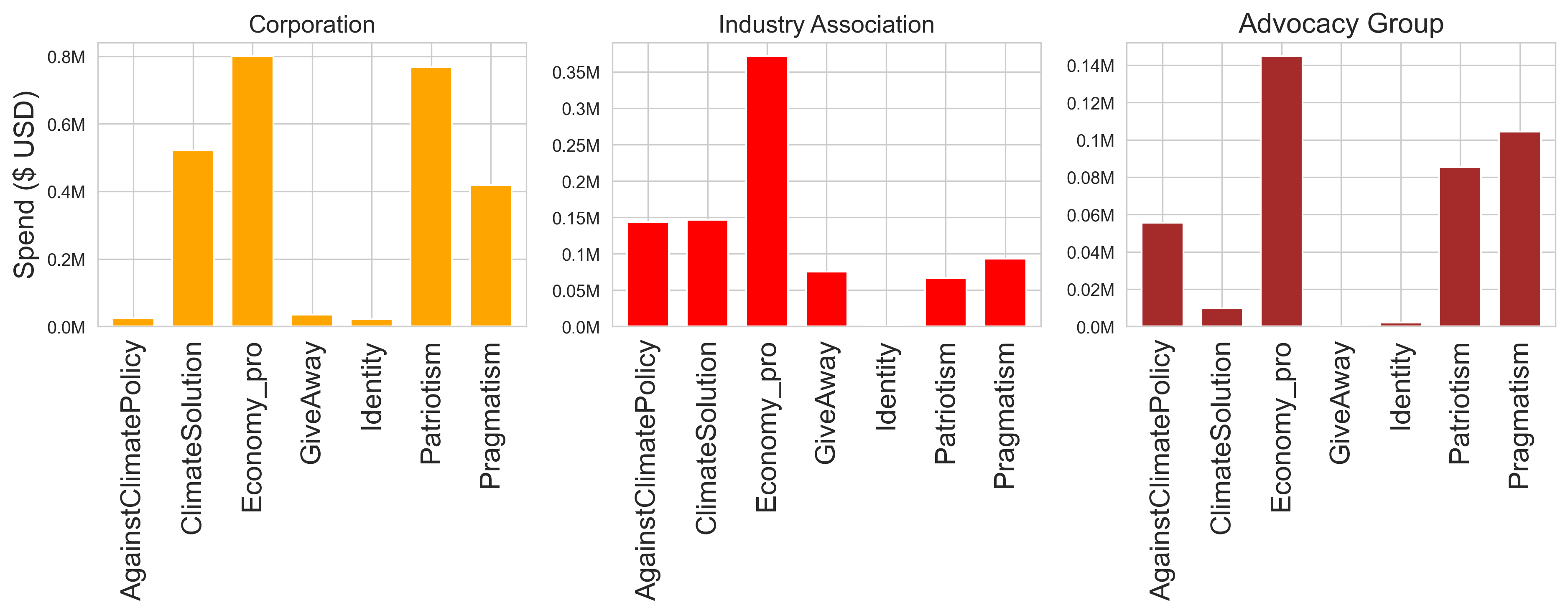}
\caption{Pro-energy ad themes by funding entity type.}
    \label{fig:thm_fe}
\end{figure*}
\subsection{Distribution of Messaging by Entity Type}
\label{fe_type}
Fig. \ref{fig:cost} shows the top $5$ funding entities based on expenditure in pro-energy and clean-energy ads. We notice that \textbf{Exxon Mobil Corporation}, which is one of the world's largest publicly traded international oil and gas companies\footnote{\url{https://corporate.exxonmobil.com/}}, spends the most on sponsoring pro-energy ads on Facebook. Clean-energy ads are mostly sponsored by \textbf{The Climate Pledge}, which is powered by $378$ companies in $34$ countries around the globe\footnote{\url{https://www.theclimatepledge.com/}}.

% \begin{table}[]
%     \centering
%     \resizebox{1\columnwidth}{!}{%
%     \begin{tabular}{|l|l|}
%       \hline
%       \thead{\textbf{Type}} & \thead{\textbf{Entity}} \\
%       \hline
%       \textbf{Corporation} & EXXON MOBIL CORPORATION \\
%       \textbf{Corporation} & Shell \\
%       \textbf{Corporation} & BP CORPORATION NORTH AMERICA INC. \\
%       \textbf{Corporation} & Twin Metals Minnesota \\
%       \textbf{Corporation} & Wink to Webster Pipeline LLC \\
%       \textbf{Industry Association} & AMERICAN PETROLEUM INSTITUTE \\
%       \textbf{Industry Association} & New York Propane Gas Association \\
%       \textbf{Industry Association} & Texas Oil \& Gas Association \\
%       \textbf{Industry Association} & New Mexico Oil and Gas Association \\
%       \textbf{Industry Association} & National Propane Gas Association \\
%       \textbf{Advocacy Group} & Coloradans for Responsible Energy Development \\
%       \textbf{Advocacy Group} & Grow Louisiana Coalition \\
%       \textbf{Advocacy Group} & Voices for Cooperative Power \\
%       \textbf{Advocacy Group} & Consumer Energy Alliance \\
%       \textbf{Advocacy Group} & Maine Affordable Energy \\
%     \hline
%     \end{tabular}}
%     \caption{List of entities from pro-energy ads.}
%     \label{tab:entity}
% \end{table}
To understand how fossil fuel industries and their support groups influence public opinion, we categorize pro-energy funding entities into three types, i.e., Corporations, Industry Associations, and Advocacy Groups. Finally, we select the top $5$ pro-energy funding entities based on their expenditure for each category. Table \ref{tab:entity} shows the list of pro-energy entities included in our analysis.

%Appendix \ref{app:et} shows the list of our selected entities. 
The highest spending on `\textbf{Economy\_pro}' narratives comes from all three entity types (Fig. \ref{fig:thm_fe}). Corporation entities spend on `\textbf{Patriotism}' narratives as their second target. Furthermore, advocacy groups focus on `\textbf{Pragmatism}' narratives as their second target. Moreover, industry associations spend almost equally on `\textbf{ClimateSolution}' and `\textbf{AgainstClimatePolicy}' narratives. Analyzing the messaging themes for different funding entities indicates different groups are fulfilling different messaging roles (Answer to \textbf{RQ4}). 

\section{Conclusion}
We propose a minimally supervised model soup approach leveraging messaging themes to identify stances of climate related ads on social media. To the best of our knowledge, our work is the first work that uses a probabilistic machine learning approach to analyze climate campaigns. We hope our approach of stance detection and theme analysis will help policymakers to navigate the complex world of energy. 
%Our dataset and code will be publicly available upon acceptance. 

\section{Limitations}
% This research faced limitations due to the lack of transparency of social media. They should be transparent on how
% their platforms are being used to influence the debate on the climate crisis to tackle misinformation. 
In this work, we predict the stances of ads using the theme information. We can further explore other potential tasks, such as moral foundation analysis \cite{haidt2004intuitive,haidt2007morality}, which will help  model the dependencies between the different levels of analysis. 

Note that our fine-tuned SBERT based theme assignment model is an unsupervised learning approach and an alternative approach could be zero-shot and/or few-shot classification models \cite{brown2020language}. We leave this exploration for future work.

Moreover, our analysis might have an unknown bias as it is based on English written ads on Facebook focusing on the United States only. Another limitation is transparency – some particular aspects of the
advertising campaigns are not available to the public through the Facebook Ads Library API, thus limiting our findings.

\section{Ethics Statement}
The data collected in this work was made publicly available by Facebook Ads API. The data does not contain any personally identifying information and reports engagement patterns at an aggregate level. The authors' personal views are not represented in any qualitative result we report, as it is solely an outcome derived from a machine learning model.

\section*{Acknowledgement}
We are thankful to the anonymous reviewers for their insightful comments. This work was partially supported by Purdue Graduate School Summer Research Grant (to TI) and an NSF CAREER award IIS-2048001.

%\newpage
\bibliographystyle{ACM-Reference-Format}
\bibliography{sample-base}
\end{document}